# Can neural networks do arithmetic? A survey on the elementary numerical skills of state-of-the-art deep learning models

Alberto Testolin

Department of General Psychology and Department of Mathematics
University of Padova (Italy)

alberto.testolin@unipd.it

Creating learning models that can exhibit sophisticated reasoning skills is one of the greatest challenges in deep learning research, and mathematics is rapidly becoming one of the target domains for assessing scientific progress in this direction. In the past few years there has been an explosion of neural network architectures, tasks, and benchmark data sets specifically designed to tackle mathematical problems, reporting notable success in disparate fields such as automated theorem proving, numerical integration, and discovery of new conjectures or matrix multiplication algorithms. However, despite these impressive achievements it is still unclear whether deep learning models possess an elementary understanding of quantities and symbolic numbers. In this survey we critically examine the recent literature, concluding that even state-of-the-art architectures often fall short when probed with relatively simple tasks designed to test basic numerical and arithmetic knowledge.

## 1. Introduction

Despite many animal species exhibit an approximate understanding of numbers and quantities (Dehaene, 2011; Nieder, 2016), formal mathematics is a peculiarity of *Homo Sapiens* that emerged through thousands of years of cultural evolution (Núñez, 2017; Beller et al., 2018; O'Shaughnessy et al., 2022). Mathematical reasoning requires to deploy most of our finer-grained cognitive abilities, including sophisticated pattern recognition skills, language understanding, symbolic processing, and abstract thinking, making it one of the highest achievements of human intellect. It is therefore not surprising that the scientific community has always regarded mathematical and logical reasoning as crucial steps in building intelligent machines (Newell and Simon, 1956; Bundy, 1983).

However, although computers excel at crunching numbers, solving mathematical problems remains a formidable challenge for artificial intelligence (Choi, 2021). On the one hand, grounding structured mathematical knowledge into some form of intrinsic meaning is a long-standing problem in symbolic AI (Searle, 1980; Harnad, 1990). On the other hand, neural networks always lagged in learning math, and such limitation has been traditionally considered an essential feature of their very nature, which is rooted on statistical pattern recognition abilities rather than the use of explicit syntactic rules (Fodor and Pylyshyn, 1988; Marcus, 2018). Mathematical reasoning poses well-known challenges for connectionist models: the symbols used in mathematical formulas appear as arbitrary tokens, which need to be manipulated according to well-defined rules entailing compositionality and systematicity. Furthermore, mathematical knowledge extracted from a set of examples should generalize well beyond the observed distribution, enabling extrapolation through the discovery of 'first principles'.

Despite these challenges, the recent successes of deep learning have rekindled interest in the idea that neural networks might be able to acquire high-level reasoning abilities and thus exhibit symbolic behavior (Santoro et al., 2021). Indeed, although deep networks struggle to grasp even basic concepts such as the meaning of 'integer number' (Trask et al., 2018), in the past few years several models demonstrated impressive capabilities in solving complicated mathematical tasks. For example, sequence-to-sequence architectures were shown able to learn to perform function integration and solve ordinary differential equations, sometimes with greater accuracy than popular math software packages (Lample and Charton, 2019). Deep learning models have also been successfully used for automated theorem proving (Lee et al., 2019; Polu and Sutskever, 2020; Wang and Deng, 2020) or to assist expert mathematicians to formulate conjectures and establish new fundamental results in pure mathematics (Davies et al., 2021). Last year, deep reinforcement learning discovered a more efficient algorithm to perform matrix multiplication (Fawzi et al., 2022), while fine-tuning a pre-trained language model on computer code allowed to solve university-level math questions at a human level (Drori et al., 2022).

These stunning achievements partially stem from the introduction of well-curated, large-scale data sets containing mathematical problems annotated with the corresponding solutions, but also to the design of novel (often *ad hoc*) architectures that can more effectively process numerical symbols and mathematical notation. Improvements in many tasks have also been enabled by the creation of large-scale language models, which exhibit surprising 'out of the box' numerical abilities that can be further refined through fine-tuning and prompting strategies. However, as I will highlight in the present survey, these findings do not imply that such models fully understand the semantics of numbers and basic arithmetic. In fact, their performance on relatively simple numerical tasks is often brittle, suggesting that we might need to improve the elementary skills of these models to bootstrap more reliable mathematical capabilities. This hypothesis is also supported by the extended literature on child development and education, which has shown that basic numeracy skills such as counting, quantity comparison, understanding of number ordering, and the base-ten positional numeral system are strong predictors of later mathematics achievement (Jordan et al., 2009; Claessens and Engel, 2013; Nguyen et al., 2016).

The survey is structured as follows: I will initially review the main tasks and data sets that have been proposed to train and test the elementary arithmetic abilities of deep learning models. These include numerical problems encoded using natural language ('math word problems') or simple mathematical formalism (e.g., multi-digit addition and subtraction), as well as other basic numeracy tasks. I will then present the main neural network architectures that have been proposed to solve this kind of problems, which include general-purpose deep learning models, but also *ad hoc* modules specifically tailored to process mathematical notation. I will finally review the arithmetic abilities of large language models and the main strategies that have been adopted to inject number semantics into word embeddings[1].

## 2. Elementary numerical tasks and data sets

### 2.1. Math word problems

Math Word Problems (MWPs) are rapidly becoming a standard benchmark for AI models (for a recent review, see Faldu et al., 2021). They are narrative problems that are used to assess the general understanding of numbers and operations through everyday life situations and are one of the most common types of numerical task encountered by children in primary schools. MWPs are framed in natural language and vary in complexity depending on the type of arithmetic operations they require, and whether such operations involve small or large operands.

---

[1] This survey focuses on basic numerical and arithmetic skills; for a more general overview of deep learning models for mathematical reasoning the reader could also refer to Lu et al. (2022).

Most MWPs data sets used in AI research are either curated from educational or online resources, derived by processing other available data sets, or synthetically created. One of the first large-scale data sets has been Dolphin18K (Huang et al., 2016); it contains 18,000 elementary math word problems taken from an online math repository, annotated with ground truth information using a semi-automatic procedure. A similar data set is Math23K (Wang et al., 2017), which contains 23,161 MWPs in Chinese language crawled from online education websites. A step forward in data set creation was implemented in the AQuA data set (Ling et al., 2017), which contains 100,000 problems annotated not only with the correct answer, but also with the sequence of steps ('answer rationale') required to derive the solution. Problems were generated by starting from a set of seed problems, which were then modified through crowdsourcing. This approach for building a large-scale MWPs data set was further refined in MathQA (Amini et al., 2019) by introducing a more precise language to define operation programs representing the intermediate problem solution steps. It contains 37,200 problems taken from the AQuA data set, annotated with the corresponding lists of multiple-choice options and aligned operation programs, again using a crowdsourcing platform. An example of such problems is given in Fig. 1A.

In most of the cases discussed above, the authors implemented some control procedures to limit the possibility that crowdsources could create duplicate versions of the same problem (e.g., by copying and pasting online problems or by proposing trivial modifications to the problem text). However, the research community recently pointed out that these data sets actually contain many problems with overlapping structure or very similar content (Miao et al., 2020). This makes a large fraction of MWPs solvable through simple heuristics, for example by treating the text as bag-of-words or even ignoring the question during the generation of the answer, leading to a significant overestimation of model performance (Patel et al., 2021). These shortcomings called for the design of more controlled data sets. One of them is ASDiv (Miao et al., 2020), which contains only 2,305 problem instances which, however, are provably more heterogeneous compared to previous corpora thanks to the use of a metric introduced to measure the lexicon usage diversity. This design principle was also adopted for the creation of a larger-scale benchmark called GSM8K (Cobbe et al., 2021), which contains 8,500 high quality math problems at the elementary school level, created by human problem writers. This data set was similarly built with the goal of featuring high linguistic diversity, while relying on relatively simple math concepts: problems take between 2 and 8 steps to solve, and solutions primarily involve performing a sequence of elementary calculations using basic arithmetic operations. Another recently proposed benchmark is SVAMP (Patel et al., 2021), a data set containing 1,000 simple (one-unknown) arithmetic word problems of grade level up to 4. It was created by applying a set of variations to existing problems sampled from the ASDiv data set (Miao et al., 2020), which were carefully designed to probe whether the model's answer actually depends on the question ('question sensitivity'), whether the model has the capacity to determine a change in reasoning arising from subtle changes in the problem text ('reasoning ability') and whether the answer remains invariant to superficial structural changes that do not alter the problem semantics ('structural invariance').

## 2.2. Problems encoded using simple mathematical notation

Other benchmarking approaches directly probe mathematical knowledge using numerical symbols and formal notation. In this case, the challenge is mostly to demonstrate that the model can extrapolate beyond the range of numbers and operations encountered during training, for example by counting a greater number of items or solving arithmetic problems with more (and possibly longer) operands.

In this setting, test problems are normally generated using synthetic procedures. One landmark work (Trask et al., 2018) proposed to evaluate neural networks using simple function learning tasks (i.e., arithmetic operations), in some cases requiring an initial step of perceptual processing, as in the 'MNIST Digit Counting' and 'MNIST Digit Addition' tasks. Model performance is then assessed on held-out values from within the training range (interpolation) or from outside of the training range (extrapolation). A similar procedure was also adopted in more recent studies, with the goal of systematically characterizing extrapolation capabilities across numerical ranges (Madsen and Johansen, 2020; Cognolato and Testolin, 2022; Fujisawa and Kanai, 2022) or to investigate 'length generalization' as the ability to extrapolate from short problem instances to longer ones (Anil et al., 2022). Another class of basic tasks that are particularly challenging in the extrapolation regimen involves the translation from number words to quantities, as in the 'language to number translation task' (Trask et al., 2018) or, *vice versa*, from quantities to number words. These tasks probe whether the compositional structure of number words and number symbols is learned in a systematic way. An example of such problems is given in Fig. 1B.

One of the most popular synthetic benchmarks containing math problems encoded using formal notation is the Mathematics data set (Saxton et al., 2019). It was built with the main goal of providing a controlled set of problems, for example by making sure that test problems contained questions that were never seen during training. The generation method allows to classify problems according to math domain, difficulty, and the need to perform interpolation or extrapolation. The data set contains 2 million free-form questions/answer pairs, encoded as sequences of characters, spanning problems in algebra, arithmetic, calculus, number comparison, measurement, and probability. According to the authors, extrapolation abilities can be measured along different axes, for example by introducing problems involving larger numbers, more numbers, and more compositions of basic operations. Examples of such problems are given in Fig. 1C.

### 2.3. Higher-level math problems

Questions requiring numerical reasoning have also recently been included in general 'reading comprehension' benchmarks (Dua et al., 2019; Lin et al., 2020). Compared to MPWs, this kind of benchmark usually contains much longer language contexts, involves more open domain questions, and requires deeper paragraph understanding, for example, by entailing numerical reasoning over dates or ordering of events in the text paragraph. An example of such problems is given in Fig. 1D. One of the most challenging benchmarks of this kind is NumGLUE (Mishra et al., 2022b), which contains approximately 100,000 problems covering eight different types of tasks, all of which require simple arithmetic understanding. Some problems are self-contained, while others require additional background knowledge to produce the final solution, such as commonsense reasoning (e.g., *"How many faces do 10 dice have?"*).

Another recently introduced data set is MATH (Hendrycks et al., 2021), which consists of 12,500 problems taken from high-school math competitions that span a variety of math domains, carefully annotated with step-by-step solutions. Given the difficulty of these problems, the authors also released a large-scale pretraining data set called Auxiliary Mathematics Problems and Solutions (AMPS), which contains over 100,000 Khan Academy problems with step-by-step solutions in Latex format; these exercises are used to teach human students concepts ranging from elementary math to multivariate calculus.

The authors of NumGLUE also recently proposed LILA (Mishra et al., 2022a), a 'unified' mathematical reasoning benchmark consisting of 23 mathematical reasoning tasks. It has been built by extending 20 existing data sets spanning a wide range of topics in mathematics, matched (in a semi-automatic way) with corresponding Python programs that serve as reasoning chains for each question in the benchmark. The authors also include an additional benchmark data set to measure performance specifically on out-of-distribution examples and to test model robustness to language perturbations, similar to Patel et al. (2021).

Overall, the performance level on these more challenging benchmarks is far from ceiling and neural network models perform significantly worse than humans (Lewkowycz et al., 2022; Frieder et al., 2023). However, it should be noted that these data sets are challenging also for humans: university students have been estimated to reach around 40% on MATH, while a three-time IMO gold medalist attained 90% (Hendrycks et al., 2021).

```
A   Problem:           Bruce purchased 7 kg of grapes at the rate of 70 per kg and 9 kg of
                       mangoes at the rate of 55 per kg. How much amount did he pay to the
                       shopkeeper?
    Operation stack:   a = multiply(7, 70); b = multiply(9, 55); add(a,b)
    Answer:            985

B   Problem:           Write the Arabic number corresponding to the following number words:
                       "three hundred and thirty-four"
                       "one thousand two hundred and seven"
                       "eighty-nine thousand and one"
    Answer:            334 1207 89001

C   Problem:           Divide 1136975704 by -142121963.
    Answer:            -8

    Problem:           Let k(u) = u**2+u-4. Find k(0).
    Answer:            -4

D   Problem:           Before the UNPROFOR fully deployed, the HV clashed with an armed force
                       of the RSK in the village of Nos Kalik, located in a pink zone near
                       Sibenik, and captured the village at 4:45 p.m. on 2 March 1992. The JNA
                       formed a battlegroup to counterattack the next day.
    Question:          What date did the JNA form a battlegroup to counterattack after the
                       village of Nos Kalik was captured?
    Answer:            3 March 1992
```

**Figure 1**. Examples of problems from representative data sets. A) MathQA (Amini et al., 2019). B) Language to number translation task (Trask et al., 2018). C) Mathematics (Saxton et al., 2019). D) NumGLUE (Mishra et al., 2022b).

## 3. Neural network models for numerical reasoning

### 3.1. Generic deep learning architectures

Over the years, researchers have attacked numerical reasoning tasks using a variety of neural network approaches. Most of the initial work was based on generic architectures, such as long-short term memory networks (Hochreiter and Schmidhuber, 1997) and sequence-to-sequence models (Sutskever et al., 2014). The rationale for using a generic architecture is that learning to perform numerical reasoning might not be qualitatively different from the acquisition of domain-general 'abstract reasoning' skills and should thus be treated as a general (though particularly challenging) learning problem. However, it has been repeatedly pointed out that neural networks exhibit poor numerical extrapolation capabilities (Trask et al., 2018), and sequence-to-sequence or even advanced transformer-based architectures often fail in tasks requiring several intermediate calculations, where humans can instead easily find a solution (Saxton et al., 2019). The related question of whether and how recurrent networks generalize to sequences longer than those encountered during training has also been of enduring interest, and definitive solutions to such challenging issue have not yet been found (Anil et al., 2022).

The research community has been actively exploring novel domain-general neural architectures that might overcome these limitations. One interesting approach to tackle algorithmic learning (a.k.a. 'program synthesis') has been to augment neural networks with external memory modules, which enable systematic abstraction even from few examples and can (at least theoretically) learn to approximate any algorithmic procedure (Graves et al., 2016; Santoro et al., 2016). Interestingly, neural models equipped with external memory have indeed shown able to generalize well beyond their training range in binary addition and multiplication problems (Kaiser and Sutskever, 2015). Subsequent work has further refined this results by improving the format of the external memory, for example by showing that a grid-like memory representation can significantly improve out-of-distribution generalization on decimal addition, both in terms of digit length and number of operands (Kim et al., 2021b; Cognolato and Testolin, 2022).

Interestingly, the idea of granting access to an external memory to solve complex problems is reminiscent of the notion of 'material representation' introduced in anthropology, which has been recently elaborated in the context of numerical cognition (Overmann, 2016). According to this view, abstract mathematical concepts would be a relatively recent cultural achievement, which emerged thanks to the spread of numerical manipulation tools (d'Errico et al., 2018). This perspective has recently been explored in computational models based on deep reinforcement learning, which can simulate the active interaction of a learning agent with external numerical representation devices (Sabathiel et al., 2022; Petruzzellis et al., 2023). A related stream of research seeks to improve large-scale models by granting them access to external tools (Parisi et al., 2022) or by constructing modular architectures equipped with discrete knowledge bases and reasoning components (Karpas et al., 2022).

### 3.2. *Ad hoc* deep learning architectures

Given the difficulty of building generic deep learning architectures that can achieve algorithmic generalization, many authors have instead focused on creating 'neuro-symbolic' systems that combine neural networks with rule-based numerical solvers, or to design neural network modules specifically tailored for numerical reasoning.

One of the earliest attempts to solve math word problems using a neuro-symbolic approach consisted of a vanilla sequence-to-sequence model combined with a similarity-based retrieval model, with the goal of creating a hybrid MWPs solver (Wang et al., 2017). Several subsequent approaches exploited graph-based problem representations, which facilitate the solution of MWPs through the use of structured descriptions reflecting a tree-like algorithmic decomposition of the problem (Wang et al., 2018; Xie and Sun, 2019) or explicitly capturing the relationship between quantities and their attributes (Zhang et al., 2020a). Another recent *ad hoc* architecture incorporates numerical reasoning into the system by exploiting a 'numerically-aware' graph neural network (Ran et al., 2019): the model is composed by standard neural encoding and decoding modules, but also includes a 'reasoning module' that represents the MWP quantities using a graph, where the nodes correspond to the numbers appearing in the text and the edges encode magnitude relationships (i.e., 'greater than'). However, since the graph is pre-defined for each problem, such model cannot deal with tasks entailing the generation of intermediate numerical results and has limited arithmetic reasoning capabilities. Overall, we can argue that models incorporating *ad hoc* graph-based representations generally improve over standard seq2seq architectures, but their performance is still poor when tested using more controlled (though relatively simple) data sets (Miao et al., 2020; Patel et al., 2021).

A modular design was also adopted by a recent neuro-symbolic model developed to tackle reading comprehension problems involving numerical reasoning (Dua et al., 2019). In this architecture, a transformer-based module first predicts whether the problem answer is a count or an arithmetic expression, and then identifies the specific numbers involved in the expression. Once a proper arithmetic expression has been formed, it is given as input to a symbolic solver to produce the final answer. The model improved over the considered baselines; however, its testing accuracy was still far from the human level in the data set considered by the authors (44% *vs*. 94%).

Inspired by the fact that neural networks fail to learn to represent numbers outside of the range seen during training, others proposed to augment standard artificial neurons with *ad hoc* modules biased to learn systematic numerical computation. For example, the 'Neural Arithmetic Logic Unit' is augmented with operators that can represent simple functions such as addition, multiplication, and power functions (Trask et al., 2018). The 'Neural Arithmetic Unit' generalizes this idea and achieves higher extrapolation capabilities by introducing a simplification of the parameter matrix, a sparsity regularizer, and a new multiplication unit (Madsen and Johansen, 2020). In general, these models demonstrate some extrapolation capabilities in simple arithmetic tasks, for example by accumulating moderate test errors on additions involving two 1000-digit numbers when being trained only with 10-digit numbers. However, accuracy is still far from ceiling, and these models do not easily generalize to problems involving a higher number of operands.

### 3.3. Large language models

One of the most exciting discoveries of the past few years has been to realize that large language models (LLMs) trained in an autoregressive way can exhibit a surprising level of competence in a variety of tasks 'out of the box', that is, without having been explicitly trained to solve such problems. This seems to be the case also for numerical reasoning: for example, GPT-3 is able to carry out basic arithmetic tasks such as two-digit addition without any additional fine tuning, with performance becoming progressively stronger moving from the zero-shot to one-shot to few-shot setting (Brown et al., 2020).

However, it turns out that the numerical knowledge of LLMs is often superficial and inaccurate: the calculation accuracy of GPT-3 rapidly decreases as the number of digits increases (Brown et al., 2020) and further growing the model size does not necessarily improve extrapolation capabilities (Henighan et al., 2020; Rae et al., 2021); the text-to-text T5 model fails in simple numerical tasks, such as number comparison and number sorting, when probed outside the interpolation range (Pal and Baral, 2021); the mathematical abilities of the popular ChatGPT model are significantly below those of an average mathematics graduate student (Frieder et al., 2023). In general, even the largest models falter when required to perform multi-step mathematical reasoning, such as those included in the MATH benchmark. It has been shown that the performance of LLMs on mathematical calculations strongly correlates with term frequency in the training data (Razeghi et al., 2022), suggesting that these gigantic models might obtain seemingly high performance from surface-level memorization, rather than understanding of arithmetic procedures.

One possibility to improve the numerical reasoning in LLMs is to fine-tune them using domain-specific data sets. For example, GPT-style models can solve many of the problems in the Mathematics benchmark once trained on math data sets (Henighan et al., 2020), and fine-tuned BERT-style and T5-style models achieve significant improvements also on more challenging benchmarks (Geva et al., 2020; Yang et al., 2021). Yet, these LLMs still fail even in simple numerical tasks that probe extrapolation capabilities, and sometimes performance is even worse than the original architectures (Pal and Baral, 2021).

Performance can also be improved by combining generative LLMs with post-processing verifiers, which are trained to judge the correctness of model-generated solutions: at test time, a fixed number of candidate solutions is sampled and the one ranked highest by the verifier is selected as final output (Cobbe et al., 2021). In this specific case, the system was also further trained to rely on a calculator for solving arithmetic problems, by injecting calculation annotations into the training set; at test time, the calculator overrides sampling when the model produces the annotations. The idea of producing annotations that can be subsequently processed by a software calculator is also pursued in more sophisticated approaches, which train LLMs to generate computer code to solve complicated algorithmic tasks, rather than directly asking for the solution. In one recent demonstration of this method (Drori et al., 2022), the authors exploited an LLMs pre-trained on text and fine-tuned on computer code (Chen et al., 2021) to generate computer programs that solve challenging (university-level) problems from the MATH benchmark. The performance gain was significant, though it should be emphasized that in these cases the task of producing the final solution is partially delegated to an external (e.g., Python) interpreter, which can also take advantage of external libraries to perform mathematical operations such as solving equations or taking limits.

*Promoting step-by-step numerical reasoning*

Another promising approach to improve the numerical abilities of LLMs involves the use of advanced *prompting* strategies (a.k.a. 'in-context learning'), which allow shaping the model's behavior by providing a few input–output exemplars demonstrating the task, without actually tuning any model parameter (Brown et al., 2020). A recent line of work has shown that prompting strategies can be improved using 'scratchpads' (Nye et al., 2021) and 'chain-of-thought reasoning' (Wei et al., 2022), which allow to significantly increase accuracy on multi-step mathematical reasoning problems. The idea is that rather than having to generate a final answer immediately, the model can first generate solutions that may contain intermediate computations (see examples in Fig. 2).

To achieve this, the scratchpad technique introduced by Nye et al. (2021) allows the model to produce an arbitrary sequence of intermediate tokens that are stored in a buffer (the scratchpad) and that can be further processed before producing the final answer. The authors considered the task of learning long integer addition, showing that the use of scratchpads improves the calculation accuracy in the extrapolation interval. This idea was taken further by Wei et al. (2022): the chain-of-thought is a series of intermediate natural language reasoning steps that are provided as input to the model during prompting to explicitly demonstrate how to produce the final output. This technique was tested in arithmetic reasoning tasks, achieving striking performance gains compared to baseline models. Interestingly, it turns out that chain-of-thought prompting has larger performance gains for more complicated problems, and mostly when it is applied to larger-scale models. At the time of publication, it established a new state-of-the-art performance on the challenging GSM8K and SVAMP benchmarks.

Similarly to what happens in recurrent models with adaptive computation time (Graves, 2016; Banino et al., 2021), these advanced prompting techniques allow the neural network to process the input information for as long as needed, depending on the complexity of the current problem. By encouraging the model to produce an explanation along with the answer, we also steer it towards solving problems by breaking them into smaller steps that logically follow from each other. Furthermore, the buffer allows to store intermediate information for an arbitrary amount of processing steps, removing the need to memorize all intermediate states in the network's activations. Last, but not least, this approach provides an interpretable window into the behavior of the model, suggesting how it might have arrived at a particular answer and providing opportunities to debug where the reasoning path went wrong. In fact, structured prompting techniques are reminiscent of the educational process adopted by teachers in primary schools, where the elementary steps for solving algorithmic problems (such as carrying out long additions) are explicitly written out to facilitate understanding.

```
A   Problem:        29 + 57
    Scratchpad:     2 9 + 5 7 , C: 0
                    2 + 5 , 6 C: 1 # added 9 + 7 = 6 carry 1
                    , 8 6 C: 0 # added 2 + 5 + 1 = 8 carry 0
                    0 8 6
    Answer:         86
```

```
B   Problem:        If there are 3 cars in the parking lot and 2 more cars arrive, how many
                    cars are in the parking lot?
    Chain-of-thought: There are originally 3 cars. 2 more cars arrive. 3 + 2 = 5.
    Answer:         The answer is 5.

    Problem:        There were nine computers in the server room. Five more computers were
                    installed each day, from monday to thursday. How many computers are now
                    in the server room?
    Chain-of-thought: There were originally 9 computers. For each of 4 days, 5 more computers
                    were added. So 5 * 4 = 20 computers were added. 9 + 20 is 29.
    Answer:         The answer is 29.
```

**Figure 2**. Examples of numerical and math word problems solved using advanced prompting strategies. A) The scratchpad method forces the model to explicitly produce intermediate computation steps, which are iteratively refined to generate the final answer (Nye et al., 2021). B) A subset of chain-of-though prompting exemplars that were given as input to the model to elicit step-by-step reasoning during problem solution (Wei et al., 2022).

These methods have only scratched the surface of the potential of prompting for eliciting high-level reasoning. One of the most advanced LLMs designed for solving mathematical reasoning problems is currently Minerva (Lewkowycz et al., 2022), which is based on a version of the PaLM language model that was fine-tuned on a high-quality dataset containing scientific and mathematical data, specifically built by crawling arXiv papers and web pages containing math formulas. The model exploits chain-of-thought prompting, and also generates multiple candidate solutions that are selected using a majority voting scheme. Minerva significantly outperforms the original PaLM model and established a new state-of-the-art performance on the GSM8K and SVAMP benchmarks, although its average accuracy remains below the human level.

Another recent improvement has been the introduction of algorithmic prompting methods (Zhou et al., 2022), which involves providing a detailed description of the algorithm execution on running examples and using explicit explanations and natural language instructions to further remove ambiguity. This allows to drastically increase the amount of detail included in the input rationales, at the same time promoting the composition of basic skills to solve complex reasoning tasks. Such a prompting strategy outperforms existing prompting techniques on several algorithmic tasks, including arithmetic, and achieves much higher performance on addition problems, also in the extrapolation range.

### *Injecting numerical semantics into word embeddings*

It turns out that the way numbers are represented in their 'surface form' has a strong influence on the model performance in numerical reasoning tasks. Although some initial investigations suggested that standard embedding techniques could capture number semantic fairly well (Wallace et al., 2019), subsequent studies pointed out that these representations are in fact inadequate at dealing precisely with numbers (Naik et al., 2019) and that commonly used sub-word tokenization techniques disrupt magnitude information (Nogueira et al., 2021). For example, in GPT-3 a number like 1598 is tokenized as '15' and '98', while another format like 1,598 is split as three different tokens: '1', ',', and '598'.

Simple tricks to improve model performance are to adopt a character-level encoding of numbers, or to replace numbers with their corresponding scientific notation (Zhang et al., 2020b). Another effective workaround consists of using more meaningful surface representations, for example by explicitly providing the 10-base number decomposition (e.g., '832' becomes '8 100 3 10 2') (Kim et al., 2021a; Nogueira et al., 2021). More advanced approaches have tried to augment the standard embeddings of number words by explicitly injecting semantic information representing quantities (for a survey, see Thawani et al., 2021). One of the first attempts at learning better numeral embeddings (Jiang et al., 2020) proposed to first map numbers into a log-space (to compress larger values) and train either a self-organizing map or a Gaussian mixture model with the goal of creating latent vectors encoding 'number prototypes'. A similarity function is then used to project the input number into the closest prototype. Another method forced the cosine similarity between word embeddings of two numbers to reflect their actual distance on the number line (Sundararaman et al., 2020). Further refinements combine the prototype-based approach with the scientific notation encoding (Jin et al., 2021). Another interesting technique exploits the Online Encyclopedia of Integer Sequences, which includes notable series of numbers that are of particular mathematical interest, to train more meaningful number embeddings (Ryskina and Knight, 2021). The rationale was to learn number embeddings by looking at number co-occurrence across well-structured, regular number sequences, to see if this would lead to the emergence of encodings representing useful relational properties of the integers. For example, the authors discovered that one specific neuron in the embedding vector was positively activated for even numbers and negatively activated for odd numbers, suggesting the emergence of a localistic encoding of an 'evenness' feature.

Overall, all these approaches for augmenting numerical embeddings allow to improve processing of small and medium-sized numbers, especially on linguistic tasks such as math word problems; however, none of them allows to perform accurate manipulation (e.g., arithmetic) over larger quantities or numbers that are not contained in the training vocabulary.

## 4. Conclusion

To summarize the main findings from the recent literature, we can generally argue that although neural network models can exhibit impressive mathematical reasoning skills, their basic abilities for representing and manipulating numbers are still unsatisfactory. Even the most advanced deep learning architectures, including large language models, often fail when probed with numerical problems that are posed in a different manner than the training cases or that involve quantities well outside the training distribution.

The state-of-the-art performance on GSM8K is still far from the human level, ranging from 58% for PaLM with 540 billion parameters, chain-of-thought prompting and access to an external calculator (Chowdhery et al., 2022) to 78% for Minerva (Lewkowycz et al., 2022). The top accuracy on ASDiv and SVAMP is higher, achieving almost 90% for large language models with refined chain-of-thought prompting strategies (Wang et al., 2022). However, achieving high performance on these benchmarks might not be enough to demonstrate numerical understanding, as also pointed out by others (Davis, 2023). Not surprisingly, indeed, the extrapolation capabilities of these gigantic neural networks are still poor: when models are not equipped with external calculators, generalization to unseen (out-of-distribution) numerical ranges requires the discovery of algorithmic procedures, which still constitutes a formidable challenge for deep learning (Welleck et al., 2022) and is indeed considered one of the frontiers in neuro-symbolic research (Hitzler and Sarker, 2021). For example, Minerva achieves over 80% accuracy on 10-digit addition and over 20% accuracy on 18-digit addition, but calculation accuracy is never at ceiling even for easy (e.g., 5-digit) problems, which highlights that even such a sophisticated model still has a very limited understanding of elementary arithmetic. A similar conclusion applies to the extrapolation results presented in Zhou et al. (2022): a finer-grained algorithmic prompting strategy allows to consistently solve addition problems having answers of up to 19 digits in length, even if training examples were limited to 5 digits. However, the accuracy never reaches 100% and after 19 digits the model runs out of context.

As others advocate, I believe that building deep learning systems that can more reliably master basic numerical concepts and simple arithmetic should be a mandatory first step towards successfully tackling complex mathematical reasoning tasks (Mishra et al., 2022b). I further argue that the perspective offered by decades of research in educational, developmental, and cognitive sciences could provide important insights for the design of such learning systems, especially in terms of the design of training regimens and testing procedures. For example, primary education in mathematics places a strong emphasis on the acquisition of elementary algorithms as the basis for understanding the arithmetic of natural numbers, integers, fractions, and decimals, using a variety of representation formats that exemplify the semantics of the decimal place-value system (Sarama and Clements, 2009). Importantly, such knowledge is developed on top of the acquisition of number words, numeral systems and counting procedures (Rittle-Johnson et al., 2001; Carey and Barner, 2019), which are fostered even during the pre-school period (Anders et al., 2012). Furthermore, although language has a central role in this arduous learning process, it is not the only medium through which numerical knowledge is acquired (Gelman and Butterworth, 2005): a key role is also played by the development of basic visuo-spatial perceptual skills and geometrical intuitions (Dehaene, 2009; Kellman et al., 2010; Piazza, 2010), as well as by the mastery of embodied and material representational systems (Lakoff and Núñez, 2000; Bender and Beller, 2012; Overmann, 2016). Cognitive science also provides well-established operational definitions and testing procedures to assess the many facets of numerical understanding (Delazer et al., 2003; Clements et al., 2008; Purpura and Lonigan, 2015); the adoption of similar evaluation batteries in AI research would allow to better characterize the numerical skills of deep learning models and to more systematically identify their weaknesses.

At the same time, implementing computational models that more faithfully simulate the acquisition of basic numerical skills would constitute an important step toward improving the scientific understanding of our own mathematical learning abilities (Testolin, 2020). Deep learning has already provided key insights into the origin of our 'number sense', for example by demonstrating that approximate numerical representations can emerge in a variety of generative architectures (Stoianov and Zorzi, 2012; Zhao et al., 2018; Testolin et al., 2020a; Boccato et al., 2021) and that number acuity becomes gradually refined through unsupervised learning (Testolin et al., 2020b). Although these modeling efforts have not yet been successfully extended into the realm of symbolic mathematics, the rapid progress in artificial intelligence is opening exciting prospects for bridging this gap.